\documentclass[10pt,twocolumn,letterpaper]{article}

 \usepackage[pagenumbers]{cvpr} % To force page numbers, \eg for an arXiv version
\usepackage{multirow,bigstrut}

\definecolor{cvprblue}{rgb}{0.21,0.49,0.74}
\usepackage[pagebackref,breaklinks,colorlinks,allcolors=cvprblue]{hyperref}
\newcommand{\bfsection}[1]{\vspace*{0.00cm}\noindent\textbf{#1.}}

\title{Hipandas: Hyperspectral Image Joint Denoising and Super-Resolution \\
by Image Fusion with the Panchromatic Image}

\author{Shuang Xu$^{1,2}$, Zixiang Zhao$^{3}$, Haowen Bai$^{4}$, Chang Yu$^{4}$, Jiangjun Peng$^{1,2}$, Xiangyong Cao$^{4}$, Deyu Meng$^{4}$\\
	$^1$ Northwestern Polytechnical University, \\
	$^2$ Research \& Development Institute of Northwestern Polytechnical University in Shenzhen,	\\
	$^3$ ETH Z\"urich, $^4$ Xi'an Jiaotong University
}

\begin{document}
\maketitle

\begin{abstract}
Hyperspectral images (HSIs) are frequently noisy and of low resolution due to the constraints of imaging devices. Recently launched satellites can concurrently acquire HSIs and panchromatic (PAN) images, enabling the restoration of HSIs to generate clean and high-resolution imagery through fusing PAN images for denoising and super-resolution. However, previous studies treated these two tasks as independent processes, resulting in accumulated errors. This paper introduces \textbf{H}yperspectral \textbf{I}mage Joint \textbf{Pand}enoising \textbf{a}nd Pan\textbf{s}harpening (Hipandas), a novel learning paradigm that reconstructs HRHS images from noisy low-resolution HSIs (LRHS) and high-resolution PAN images. The proposed zero-shot Hipandas framework consists of a guided denoising network, a guided super-resolution network, and a PAN reconstruction network, utilizing an HSI low-rank prior and a newly introduced detail-oriented low-rank prior. The interconnection of these networks complicates the training process, necessitating a two-stage training strategy to ensure effective training. Experimental results on both simulated and real-world datasets indicate that the proposed method surpasses state-of-the-art algorithms, yielding more accurate and visually pleasing HRHS images.
\end{abstract}

% \vspace{-1em}
\section{Introduction}
\label{sec:intro}
% \vspace{-.5em}
Hyperspectral imaging has emerged as a powerful tool in remote sensing \cite{9451654}. Nevertheless, the acquisition of hyperspectral images (HSIs) is often impeded by practical limitations. Factors such as atmospheric turbulence and limited sensor apertures commonly result in noise corruption and low spatial resolution for HSIs.

Denoising \cite{ZhangZYSYX24,ZhangDZYSZY24} and super-resolution \cite{ShangLZW24,DianSHL24,FangYKX24} have been developed to enhance the quality. In contrast to single-image restoration, the fusion of auxiliary images from other modalities potentially leads to improvement. The recent launch of satellites equipped with hyperspectral-panchromatic (HS-PAN) imaging systems, \eg the PRISMA satellite, presents a new opportunity. PAN images, distinguished by their high spatial resolution and lack of noise, serve as an ideal data source for fusion \cite{8672156,HeFLCP24}.

% provide complementary information

\begin{figure*}
    \centering
    \includegraphics[width=0.9\linewidth]{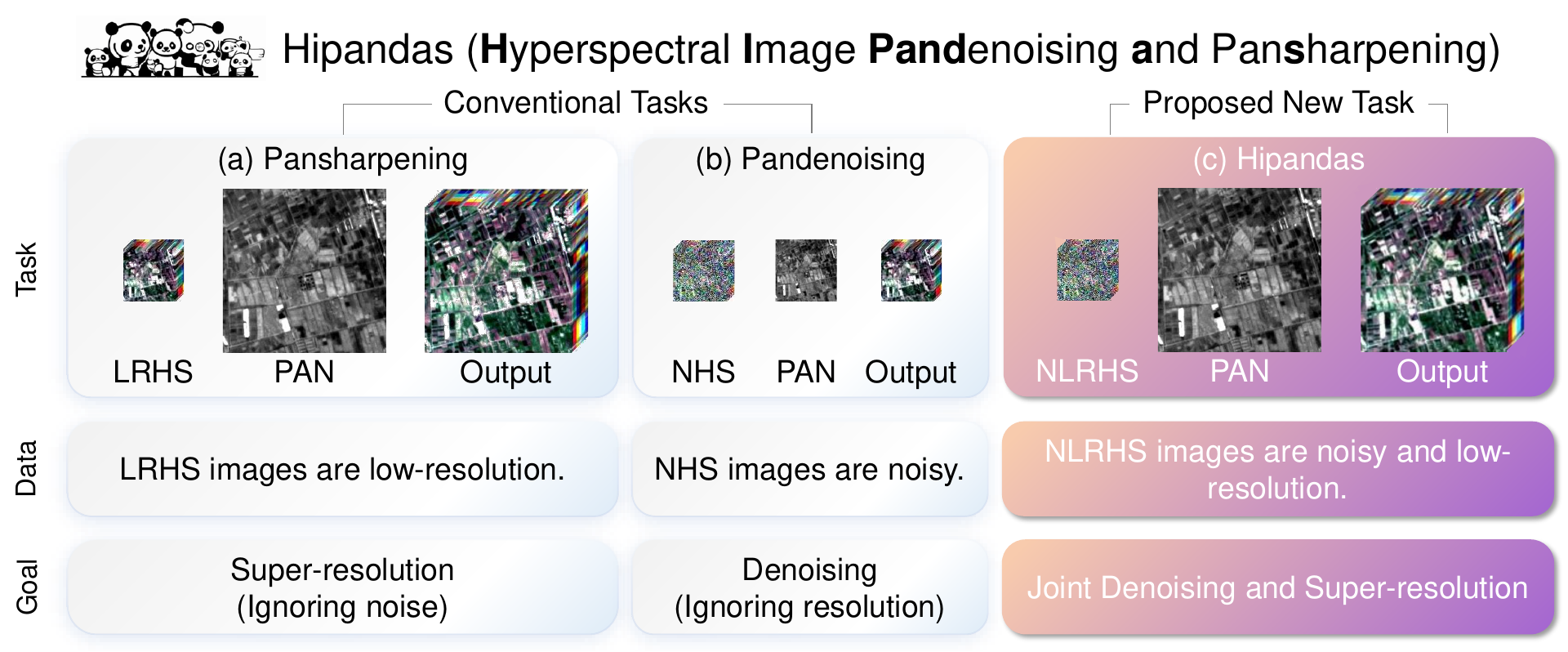}
    % \vspace{-0.5em}
    \caption{The differences among (a) pansharpening, (b) pandenoising and (c) Hipandas. Pansharpening addresses super-resolution, while pandenoising focuses on denoising. In contrast, Hipandas is specifically designed to tackle both problems in a unified framework.  }
    % \vspace{-1.5em}
    \label{fig:first_image}
\end{figure*}

As illustrated in \cref{fig:first_image}(a) and (b), guided denoising and super-resolution can be achieved through pansharpening and pandenoising techniques, respectively. However, the images captured by HS-PAN satellites (\eg PRISMA) are typically noisy and of low resolution. Currently, enhancing image quality requires sequentially applying pandenoising and pansharpening to noisy low-resolution HS (NLRHS) images. This sequential strategy poses several challenges:
\begin{itemize}
\item Denoising and super-resolution are treated as independent steps, requiring highly precise algorithms for both pandenoising and pansharpening. Inadequate denoising can lead to noise amplification during super-resolution.
\item Denoising may inadvertently smooth essential details, thereby complicating texture recovery for the super-resolution process.
\item The inherent interdependence of denoising and super-resolution implies that their disjoint execution may not fully exploit their synergistic potential.
\end{itemize}

In response to the challenges, as shown in \cref{fig:first_image}(c), the study explores a novel technique named \textit{\textbf{H}yperspectral \textbf{I}mage Joint \textbf{Pand}enoising \textbf{a}nd Pan\textbf{s}harpening (Hipandas)}, designed to reconstruct clean HRHS images from NLRHS inputs, fused with high-resolution PAN (HRPAN) images. Since real-world satellites tend to produce noisy and low-resolution HSIs, the Hipandas task is more aligned with practical scenarios than pandenoising or pansharpening tasks alone. Due to the lack of large-scale datasets for the Hipandas task, a zero-shot learning method is proposed. This method comprises a guided denoising network, a guided super-resolution network, and a PAN reconstruction network. The first two networks are utilized for image enhancement, while the third ensures modality fidelity. These networks are interconnected, with the outputs of the upstream networks serving as inputs to the downstream networks, thereby complicating the training process. Consequently, a two-stage training strategy is introduced. It is initially pretrained on low-resolution images and then finetuned on high-resolution images to mitigate learning bias towards the noise. Empirical results indicate that the two-stage training strategy overwhelmingly outperforms the one-stage strategy. In summary, the contributions are as follows:
\begin{itemize}
\item This paper proposes the Hipandas task, which is more aligned with practical scenarios.   A \textit{zero-shot learning-based Hipandas (ZSHipandas)} is formulated for this task.
\item A detail-oriented low-rank prior is introduced for the Hipandas task and is applied to the network architecture design of ZSHipandas.
\item A two-stage training strategy is proposed to effectively train ZSHipandas, which yields more accurate and visually appealing HRHS images compared to state-of-the-art (SOTA) methods.
\end{itemize}

The organization of this paper is as follows. Section \ref{sec:related_work} reviews related work. Section \ref{sec:Hipandas} delves into the detailed description of the proposed Hipandas method. Section \ref{sec:experiments} reports experimental results. Section \ref{sec:conclusion} concludes the paper.

\section{Related Work}
\label{sec:related_work}
\subsection{HSI super-resolution and pansharpening}
Before deep learning, techniques such as sparse coding \cite{BSR_HSSR,XiongZZLQ23}, wavelets \cite{PatelJ15}, and Gaussian processes \cite{AkhtarSM16} were employed for single HSI super-resolution. The success of deep RGB image super-resolution \cite{SRCNN} led to the emergence of convolutional neural networks (CNNs) as a powerful tool in the domain of single HSI super-resolution. This includes 3D CNNs \cite{LiWL21,3DFCNN}, spectral information preserving networks \cite{LiHZXL17,HuJLHZ20}, spatial-spectral prior-based networks \cite{JiangSLM20,WangMJ22}, and diffusion-based networks \cite{HSRDiff,DDS2M,WangLZ0G24,XiaoYJHJZ24}.

Despite the achievements, the fusion of additional modalities has the potential to yield better results \cite{WangDRV24}. PAN images, characterized by the high spatial resolution and signal-to-noise ratio, serve as effective guides \cite{LuZYXJ21,HyperFusion,DianGL23}. Qu \etal developed a 3D CNN to fuse HS-PAN information \cite{MSSL}. Dong \etal implemented adversarial learning for spectrum preserved HS-PAN fusion \cite{DongYQXL22}. Guarino \etal introduced a rolling pansharpening CNN (RPNN) using bandwise training to address challenges such as limited training data and variations of band number \cite{RPNN}. Li \etal proposed a dual conditional diffusion models-based fusion network that operates in an iterative denoising manner \cite{DCDMF}. Rui \etal introduced a low-rank diffusion model for hyperspectral pansharpening (PLRD), taking the strengths of a pretrained diffusion model for enhanced generalization \cite{PLRDiff}. %such as multispectral \cite{NLSTF,uSDN,FuZZZ019,YaoHCMZX20} or PAN imaging. 

\subsection{HSI denoising and pandenoising}
For HSI denoising, a range of techniques have been developed, including total variation \cite{LRTDTV,E3DTV}, nuclear norm \cite{LRMR,CTV,tCTV} and subspace methods \cite{NGMeet}. In the era of deep learning, CNNs \cite{HIDCNN,QRNN,GRN} and transformers \cite{TRQ3DNet,SST,SERT} have demonstrated promise. More recently, diffusion models have emerged as effective tools in HSI denoising. Zeng \etal utilized a transformer-based backbone for spectral unmixing, with abundance maps denoised using a self-supervised generative diffusion network \cite{DiffUnmix}. Pang \etal developed a unified HSI restoration network based on diffusion models (HIRD), which reconstructs clean HSI data from the product of two low-rank components, employing a pretrained diffusion model for base image sampling \cite{HIRDiff}. %Moreover, a novel exponential noise schedule was introduced, accelerating the restoration process by a factor of five \cite{HIRDiff}.

Extensive research has been conducted on single HSI denoising, but there is a lack of work on pandenoising. To the best of our knowledge, the study by Xu \etal \cite{PWRCTV} represents the sole study on this topic. Based on the assumption that PAN and HS images share similar textures, Xu \etal proposed a PAN-weighted total variation (PWTV) regularization \cite{PWRCTV}. Fusing PAN images enables PWTV to outperform single HSI denoising methods.

\subsection{Hipandas}
Pansharpening aims to improve spatial resolution, while pandenoising is devoted to remove noise. However, real-world satellites tend to produce noisy and low-resolution imagery, and sequential implementations of pansharpening and pandenoising is at the risk of accumulated errors. To the best of our knowledge, this paper is the early exploration on end-to-end HSI denoising and super-resolution by image fusion with PAN images.

\begin{figure*}
  \centering
    \includegraphics[width=0.9\textwidth]{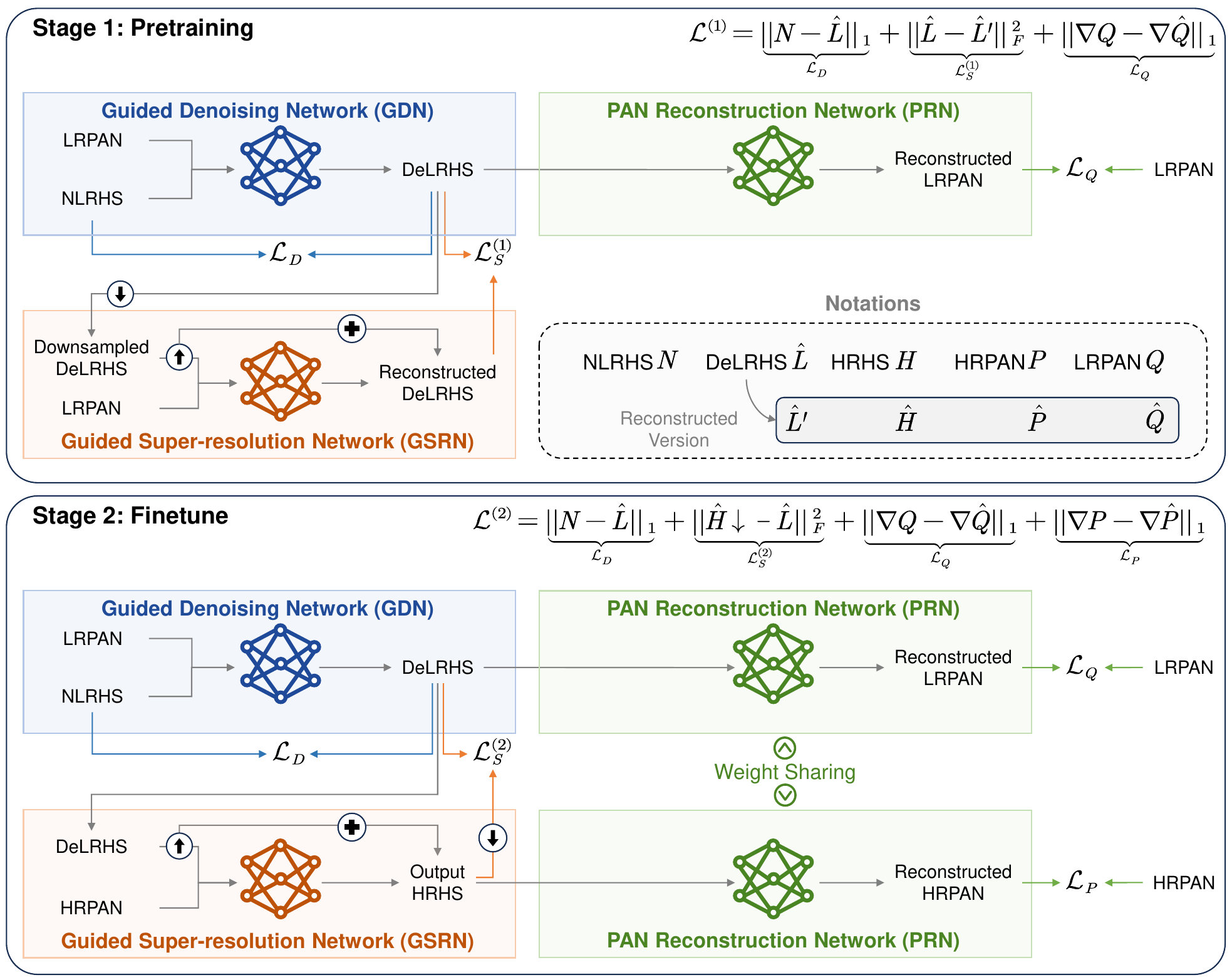}
    \vspace{-0.5em}
  \caption{The framework for the proposed ZSHipandas. It is pretrained in stage 1 on LR scale images, and then finetuned in stage 2 on HR scale images. The GDN and GSRN components are used for image enhancement, while the PRN component models relationship between HS and PAN images. }
  \label{fig:net_framework}
  \vspace{-1em}
\end{figure*}

\begin{table}
    \centering
    \caption{Main notations in this paper.}
    % \vspace{-1em}
    \resizebox{0.7\linewidth}{!}{
    \begin{tabular}{cc}
    \hline
        Variable & Denotation \\ \hline
        $P$ & High-resolution PAN (HRPAN) image\\
        $Q$ & Low-resolution PAN (LRPAN) image\\
        $H$ & High-resolution HS (HRHS) image\\
        $L$ & Low-resolution HS (LRHS) image\\
        $N$ & Noisy LRHS (NLRHS) image\\ \hline
    \end{tabular}
    }
    % \vspace{-1.5em}
    \label{tab:notations}
\end{table}

\section{Zero-Shot Hipandas}\label{sec:Hipandas} 
This section proposes the Hipandas technique, and main notations are summarized in \cref{tab:notations}.
\subsection{Problem formulation for Hipandas}
In HS imaging, the acquired data tends to be noisy and of low resolution, which can be modeled by corrupting the downsampled HRHS image with noise $\varepsilon$,
$N = H\!\downarrow + \varepsilon,$
where $N\in\mathbb{R}^{h\times w\times b}$ and $H\in\mathbb{R}^{sh\times sw\times b}$ are the noisy LRHS and HRHS images, respectively, and $\downarrow$ is the downsampling operation. Here, $s$ denotes the ratio of spatial resolution,  $h$, $w$, and $b$ denote the height, width and band number, respectively. For the HS-PAN imaging device, the system also acquires HRPAN images $P\in\mathbb{R}^{sh\times sw}$. 

PAN images, which offers complementary information, make PAN-guided HSI restoration a promising technique for achieving high-fidelity HSIs. This study introduces a method termed \textit{\textbf{H}yperspectral \textbf{I}mage Joint \textbf{Pand}enoising \textbf{a}nd Pan\textbf{s}harpening (Hipandas)}, which aims to reconstruct the noise-free HRHS image $H$ from an NLRHS image $N$ fused with the HRPAN image $P$.

\subsection{Proposed method}
Due to the limited availability of large-scale datasets with paired images $\{N,P,H\}$, this study proposes a \textit{zero-shot learning framework for Hipandas (ZSHipandas)}. As depicted in \cref{fig:net_framework}, ZSHipandas consists of three components: a Guided Denoising Network (GDN), a Guided Super-Resolution Network (GSRN), and a Panchromatic Reconstruction Network (PRN).

The GDN component is designed to eliminate noise from the LRHS image $N$, with the guidance of the PAN image to produce a denoised LRHS image $\hat{L}$. Given that the observed PAN image $P$ possesses a higher spatial resolution than $N$, the GDN component employs the LRPAN image $Q=P\!\downarrow$ to align with the spatial resolution of $N$. This process is formalized as
% \vspace{-.2em}
\begin{equation}
\hat{L}=\text{GDN}\left( N,Q \right).
\end{equation}
% \vspace{-.2em}
The GSRN component improves spatial resolution through a detail injection approach. Initially, $\hat{L}$ is upsampled (\eg, via bicubic interpolation) to generate a coarse HRHS image. Subsequently, $\hat{L}\!\uparrow$ and $P$ restore the missing details, and the final HRHS image $\hat{H}$ is expressed as
\begin{equation}\label{eq:GSRN}
\hat{H}=\text{GSRN}( \hat{L},P ) =\hat{L}\!\uparrow +f( \hat{L}\!\uparrow ,P ),
\end{equation}
where $f( \hat{L}\uparrow ,P )$ represents the predicted detail map. In addition to the aforementioned guidance during the restoration process, the strong relationship between PAN and HS images can provide valuable information. Consequently, the PRN component is employed to ensure that PAN images can be reconstructed from HSIs, formalized as
\begin{equation}
\hat{P}=\text{PRN}( \hat{H} ), \quad \hat{Q}=\text{PRN}( \hat{L} ),
\end{equation}
where $\hat{P}$ and $\hat{Q}$ denote the reconstructed PAN images.

\subsection{Loss functions and training details}
The joint training of the three networks presents specific challenges:
\begin{itemize}
\item The training data consists solely of the NLRHS image $N$, HRPAN image $P$, and LRPAN image $Q$. Lacking ground truth supervision, the formulation of rational loss functions is crucial for the training of ZSHipandas.
\item The output of the GDN component is utilized as the input for the GSRN component. However, during the early steps of training, the GDN component cannot generate a satisfactorily denoised image, which would introduce a training bias towards noise in the GSRN's training.
\end{itemize}

To mitigate these problems, loss functions tailored to the three networks are proposed, accompanied by a two-stage training strategy:

\noindent{\textbf{Stage 1.}} In this stage, the networks are pretrained at the LR scale. The reasons are twofold. LR scale images can reduce computational burden. Additionally, neural networks' spectral bias property implies that training on LR images is less complex than on HR ones \cite{xuzhiqin}. 

\noindent \textbf{a) For the GDN component}, to achieve robust denoising, an $\ell_1$ loss is utilized instead of the $\ell_2$ one between the NLRHS image $N$ and the denoised LRHS image $\hat{L}=\text{GDN}\left( N,Q \right)$, as expressed by
\begin{equation}
\mathcal{L}_{D} = \|N-\hat{L}\|_1.
\end{equation}

\noindent \textbf{b) For the PRN component}, it is imperative that the reconstructed LRPAN image maintains textures. By revising the total variation regularization, this goal is achieved by the PAN image's high-frequency information loss,
\begin{equation}
\mathcal{L}_Q=\|\nabla Q-\nabla \hat{Q}\|_1,
\end{equation}
where $\nabla$ represents the Sobel operator.

\noindent \textbf{c) For the GSRN component,} which involves super-resolution, the network cannot be trained on the observed dataset since there is no HR-LR image pair. Thus, the denoised LRHS image $\hat{L}\in\mathbb{R}^{h\times w\times b}$ is deemed as the HR data, and its downsampled version $\hat{L}\!\downarrow\in\mathbb{R}^{h/s\times w/s\times b}$ is regarded as the LR data. Then, the image could be reconstructed by $\hat{L}'=\text{GSRN}( \hat{L}\!\downarrow ,Q )\in\mathbb{R}^{h\times w\times b}$. In this way, $\hat{L}$ can be used as ground truth, and it compels the GSRN component to learn the recovery of lost details by 
\begin{equation}
\mathcal{L}_{S}^{\left( 1 \right)} = \|\hat{L}-\hat{L}'\|_{F}^{2}.
\end{equation}
In summary, the loss function in stage 1 is:
\begin{equation}
    \mathcal{L}^{(1)} = \mathcal{L}_{D}+\mathcal{L}_{S}^{\left( 1 \right)}+\mathcal{L}_Q.
\end{equation}

\noindent \textbf{Stage 2.} 
In this stage, the networks are finetuned with weights initialized from the pretraining phase. The loss functions $\mathcal{L}_{D}$ and $\mathcal{L}_Q$ remain unchanged. However, the objective of the GSRN component in this stage is to reconstruct an HRHS image, denoted as $\hat{H}=\text{GSRN}(\hat{L},P)$. In the absence of a ground truth HRHS image to supervise the learning of detail reconstruction, it has to ensure that the downsampled HRHS image retains the same content as the LRHS image. Consequently, the loss function for the GSRN component is defined as
\begin{equation}
\mathcal{L}_{S}^{\left( 2 \right)} = \|\hat{H}\!\downarrow-\hat{L}\|_{F}^{2}.
\end{equation}
Furthermore, given that the HRHS image has been reconstructed in this stage, a PAN image's high-frequency information loss is also introduced at the HR scale:
\begin{equation}
\mathcal{L}_P=\|\nabla P-\nabla \hat{P}\|_1.
\end{equation}
In summary, the loss function in stage 2 is: 
\begin{equation}
    \mathcal{L}^{(2)} = \mathcal{L}_{D}+\mathcal{L}_{S}^{\left( 2 \right)}+\mathcal{L}_Q+\mathcal{L}_P.
\end{equation}

\subsection{Network structure}
It is important to incorporate data prior into network structures. The low-rank nature of HSIs is well-established and widely applied in HSI denoising. However, the prior for HSI pansharpening is less explored. As Eq. \eqref{eq:GSRN} is based on the detail injection framework, the focus of this paper is on modeling the prior for the detail obtained by $f(\cdot, \cdot)$.

\bfsection{Low-rank Prior for detail maps}
The detail represents the information lost from an LRHS image during its reconstruction to an HRHS image. Given an HRHS image $H$, when downsampled to $L=H\!\downarrow$ and subsequently upsampled to $L\!\uparrow$, the clean detail map is derived as $D=H-L\!\uparrow$. Accounting for noise, the practical detail is the noisy version, $\tilde{D}=H-N\!\uparrow$, where $N=H\!\downarrow+\varepsilon$.

\begin{figure}
  \centering
  \begin{subfigure}{0.48\linewidth}
    \includegraphics[width=1\linewidth]{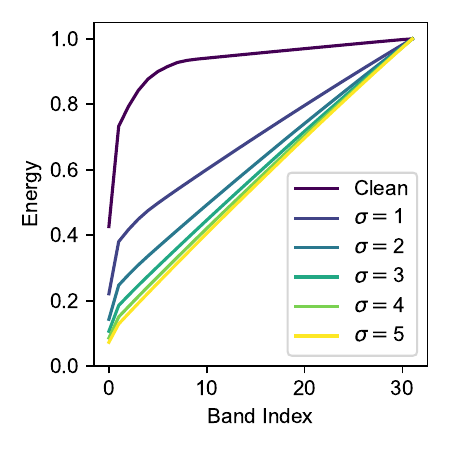}
    \caption{The Gaussian noise.}
    \label{fig:short-a}
  \end{subfigure}
  \hfill
  \begin{subfigure}{0.48\linewidth}
    \includegraphics[width=1\linewidth]{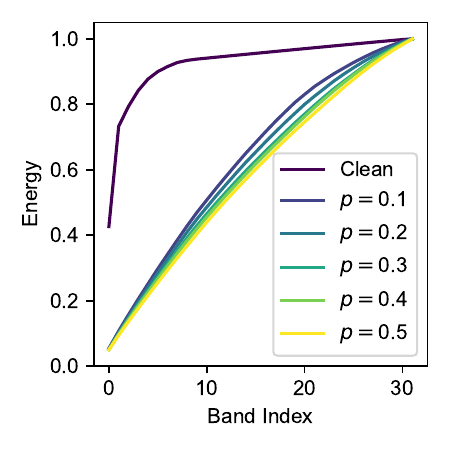}
    \caption{The mixture noise.}
    \label{fig:short-b}
  \end{subfigure}
    \vspace{-0.5em}
  \caption{The energy curve of clean and corrupted detail maps (i.e., $D$ and $\tilde{D}$). $\sigma$ and $p$ denote the noise intensity for Gaussian and mixture noise, respectively. }
  \label{fig:low_rank_detail_prior}
    \vspace{-1em}
\end{figure}

One of the main contributions of the work is the identification of the detail map $D$ as being inherently low-rank. This insight is based on an analysis of the energy curves for both clean and noisy detail maps. The $k$-th energy value is computed as $\sum_{i=1}^k \lambda_{i}/\sum_{i=1}^b \lambda_{i}$, where $\lambda_{i}$ represents the singular value of the detail map. A steeper energy curve signifies a higher degree of low-rankness in the corresponding detail map. Through simulations of noisy detail maps corrupted by Gaussian or mixed noise at varying intensities, as depicted in Fig. \ref{fig:low_rank_detail_prior}, it is observed that the clean detail map exhibits a stronger low-rank characteristic, whereas the noisy detail map tends to disrupt this property. Thus, incorporating the low-rank prior of the detail map into the network $f(\cdot,\cdot)$ facilitates the recovery of more accurate detail maps.

\begin{figure}
  \centering
    \includegraphics[width=1\linewidth]{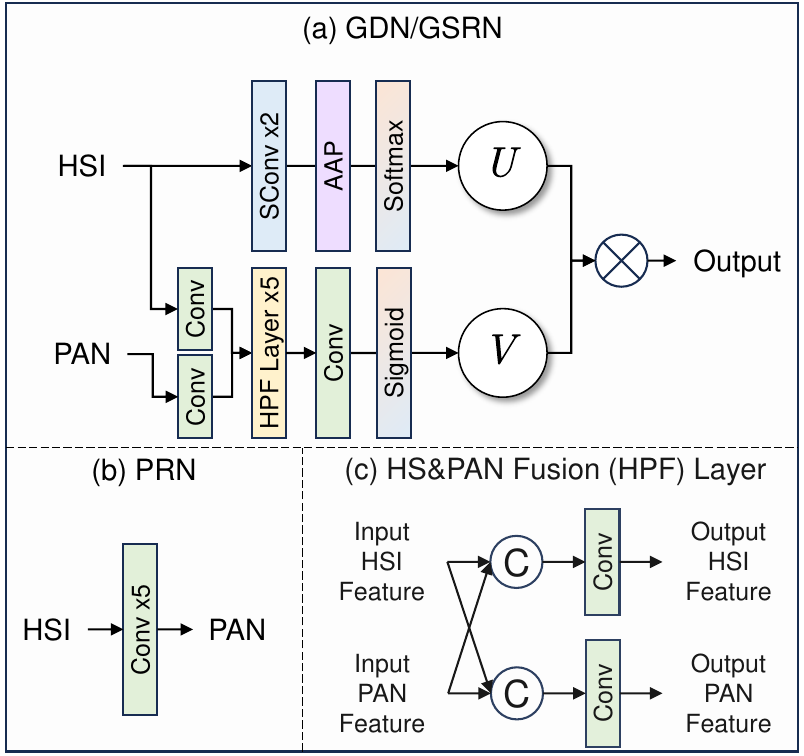}
    \vspace{-.5em}
  \caption{The architecture for the (a) GDN/GSRN component, (b) PRN component and (c) HPF layer. Conv and SConv denote convolutional and strided convolutional units, respectively. AAP denotes the adaptive average pooling. }
  \label{fig:net_structure}
    \vspace{-1em}
\end{figure}

\begin{table*}[htbp]
  \centering
  \caption{Averaged metrics on simulated data between ground truth $H$ and restored image $\hat{H}$ for the Hipandas task in Gaussian noise cases, where $\sigma$ denotes the intensity of Gaussian noise. Best and second-best values are \textbf{highlighted} and \underline{underlined}. }
  % \vspace{-1em}
  \resizebox{\textwidth}{!}{
% Table generated by Excel2LaTeX from sheet 'Sheet2'
\begin{tabular}{lcccccccccccc}
\toprule
\multicolumn{1}{c}{\multirow{2}[4]{*}{Methods}} & \multicolumn{4}{c}{i.i.d. Gaussian ($\sigma=10$)} & \multicolumn{4}{c}{i.i.d. Gaussian ($\sigma=30$)} & \multicolumn{4}{c}{non-i.i.d. Gaussian ($\sigma\in[10,50]$)} \\
\cmidrule(lr){2-5}\cmidrule(lr){6-9}\cmidrule(lr){10-13} & PSNR$\uparrow$ & SSIM$\uparrow$ & ERGAS$\downarrow$ & SAM$\downarrow$ & PSNR$\uparrow$ & SSIM$\uparrow$ & ERGAS$\downarrow$ & SAM$\downarrow$ & PSNR$\uparrow$ & SSIM$\uparrow$ & ERGAS$\downarrow$ & SAM$\downarrow$ \\
\midrule
PLRD+HIRD & 37.93 & 0.9229  & 11.97 & 4.61  & 28.40 & 0.6866  & 31.04 & 10.78 & 27.89 & 0.7499  & 30.92 & 13.37 \\
PLRD+PWTV & 34.53 & 0.7881  & 17.11 & 7.23  & 27.42 & 0.5212  & 37.73 & 17.46 & 28.84 & 0.6198  & 35.44 & 19.81 \\
SwinIR+HIRD & 29.89 & 0.8569  & 52.50 & 7.90  & 29.80 & 0.8497  & 31.98 & 8.62  & 28.10 & 0.8267  & 43.11 & 16.08 \\
SwinIR+PWTV & 27.93 & 0.6855  & 212.50 & 13.16 & 18.63 & 0.1659  & 984.41 & 44.46 & 19.60 & 0.2446  & 966.29 & 44.54 \\
RPNN+HIRD & 36.60 & 0.9280  & 12.43 & 4.53  & 32.11 & 0.8711  & 19.98 & 7.28  & 32.43 & 0.8765  & 19.68 & 8.06 \\
RPNN+PWTV & 37.19 & 0.9342  & 12.27 & 4.31  & 29.73 & 0.6570  & 27.38 & 12.85 & 31.37 & 0.7440  & 26.13 & 13.25 \\
\midrule

HIRD+PLRD & 34.36 & 0.8308 & 450.92 & 8.34  & 27.60 & 0.6303 & 569.13 & 19.14 & 26.99 & 0.5976 & 557.02 & 31.35 \\
HIRD+SwinIR & 36.24 & 0.8904 & 13.15 & 3.61  & 34.09 & 0.8700 & 16.65 & 5.40  & 32.20 & 0.8603 & 20.17 & 8.12 \\
HIRD+RPNN & 37.40 & 0.9314 & 11.95 & 4.10  & 35.06 & 0.9194 & 15.61 & 5.74  & 33.03 & 0.9072 & 18.97 & 8.22 \\
PWTV+PLRD & 36.56 & 0.9059 & 13.65 & 4.88  & 34.09 & 0.8865 & 18.21 & 7.20  & 32.48 & 0.8777 & 21.49 & 10.07 \\
PWTV+SwinIR & 37.65 & 0.9080 & 11.18 & \underline{3.09} & 34.15 & 0.8393 & 16.53 & 5.38  & 34.26 & 0.8370 & 17.71 & 6.33 \\
PWTV+RPNN & \underline{38.34} & \underline{0.9415} & \underline{10.88} & 3.87  & \underline{36.70} & \textbf{0.9314} & \underline{13.20} & \underline{4.76} & \underline{36.70} & \underline{0.9299} & \underline{13.46} & \underline{4.97} \\ 
\midrule
ZSHipandas (ours) & \textbf{40.61} & \textbf{0.9547} & \textbf{8.50} & \textbf{3.00} & \textbf{36.88} & \underline{0.9271} & \textbf{12.35} & \textbf{4.63} & \textbf{37.24} & \textbf{0.9313} & \textbf{11.87} & \textbf{4.52} \\
\bottomrule
\end{tabular}%
    }
    % \vspace{-1em}
  \label{tab:gauss}%
\end{table*}%
\begin{figure*}
    \centering
    \includegraphics[width=.95\linewidth]{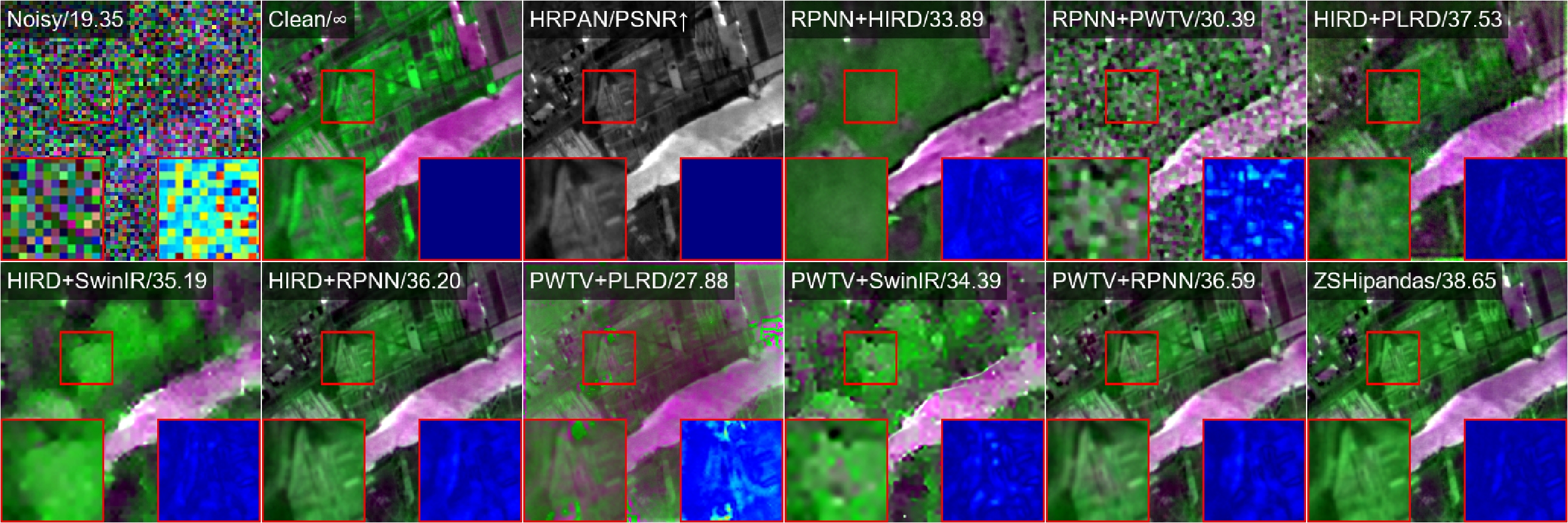}
    \vspace{-.5em}
    \caption{Restoration results for Gaussian noise with $\sigma=30$.}
    \label{fig:g30}
    \vspace{-.5em}
\end{figure*}

\bfsection{Network designing}
For GDN and GSRN components, their architectures are designed as a guided low-rank matrix factorization network, leveraging low-rank priors for both HSIs and detail maps. As depicted in Fig.~\ref{fig:net_structure}(a), the network output is represented as the product of two factor matrices. Taking the GDN component as an instance, the denoised HSI is expressed as $\hat{L}=V\times_{3} U$, where $\times_{3}$ denotes the mode-3 product, $V\in\mathbb{R}^{h\times w\times r}$ constitutes $r$ base images, and $U\in\mathbb{R}^{b\times r}$ represents the expression coefficients. Specifically, the $i$-th band of $\hat{L}$ is a linear combination of $r$ base images, given by $\hat{L}_{:,:,i}=\sum_{k=1}^{r}U_{i,k}V_{:,:,k}$. By setting $r\ll b$, the model enforces a low-rank structure on network outputs.

Conventional matrix factorization models typically address the optimization problem, $\min_{U,V} \{\mathcal{L}(Y,V\times_{3} U)+\mathcal{R}_{1}(V)+\mathcal{R}_{2}(U)\}$, where $Y$ denotes the noisy observation, and $\mathcal{L}(\cdot,\cdot)$ and $\mathcal{R}_{i}(\cdot)$ denote the loss and regularization functions, respectively. The estimations of $U$ and $V$ are computed via iterative algorithms and expressed as functions of the observed data. However, this approach is heavily reliant on manually designed objective functions. 

In contrast, this work parametrizes $V$ and $U$ as functions of the observation data via neural networks, with the forward propagation of the network approximating the computational flow of iterative algorithms by the data-driven manner, thereby potentially obviating the need for hand-crafted objective functions.

The proposed architecture comprises two distinct branches. The $U$-branch models spectral information. It applies 2 strided convolution (SConv) units to the HSI, each reducing the spatial resolution by a factor of two, resulting in a feature of size $h/4\times w/4 \times b$. An adaptive average pooling operation is applied along the spatial dimensions, followed by a softmax activation, to yield the transposed coefficient matrix $U^{T}\in\mathbb{R}^{r\times b}$. The $V$-branch is designed to generate base images. Since PAN and base images share similar structures, PAN and HS images are fused in this net. It begins with 2 head convolution (Conv) units that embed both HS and PAN images into shallow features with identical channels. Five HS and PAN fusion (HPF) layers are then utilized, with a tail Conv unit transforming the deep HS feature into $V$ with $r$ channels, activated by a sigmoid function. To mitigate computational intensity, the HPF layer is designed with a simple structure, comprising two separate Conv units applied to the concatenated HS and PAN features to produce the output features. Throughout this paper, all Conv or SConv units are activated by leaky rectified linear units, unless otherwise specified.

% \vspace{-.5em}
As for the PRN component, it is commonly postulated that there exists a spectral transform matrix $\Phi\in\mathbb{R}^{1\times B}$, which characterizes the spectral degradation between PAN-HS imaging \cite{HSRNet,GPPNN}, expressed as $P=H\times_{3}\Phi$. However, this does not hold true for numerous satellite systems. The PRN component thus is designed as 5 stacked Conv units.

% \vspace{-0.5em}
\section{Experiments}\label{sec:experiments}
% \vspace{-0.5em}

\begin{table*}[htbp]
  \centering
  \caption{Averaged metrics on simulated data between ground truth $H$ and restored image $\hat{H}$ for the Hipandas task in mixture noise cases, where $p$ denotes the intensity of mixture noise. Best and second-best values are \textbf{highlighted} and \underline{underlined}. }
  % \vspace{-1em}
  \resizebox{\textwidth}{!}{
% Table generated by Excel2LaTeX from sheet 'Sheet2'
\begin{tabular}{lcccccccccccc}
\toprule
\multicolumn{1}{c}{\multirow{2}[4]{*}{Methods}} & \multicolumn{4}{c}{Mixture ($p=0.15$)} & \multicolumn{4}{c}{Mixture ($p=0.35$)} & \multicolumn{4}{c}{Mixture ($p=0.55$)} \\
\cmidrule(lr){2-5}\cmidrule(lr){6-9}\cmidrule(lr){10-13} & PSNR$\uparrow$ & SSIM$\uparrow$ & ERGAS$\downarrow$ & SAM$\downarrow$ & PSNR$\uparrow$ & SSIM$\uparrow$ & ERGAS$\downarrow$ & SAM$\downarrow$ & PSNR$\uparrow$ & SSIM$\uparrow$ & ERGAS$\downarrow$ & SAM$\downarrow$ \\
\midrule

PLRD+HIRD & 29.38 & 0.8394  & 26.88 & 12.84 & 28.81 & 0.8275  & 28.55 & 13.04 & 28.20 & 0.7876  & \underline{34.12} & 13.80 \\
PLRD+PWTV & 30.54 & 0.7028  & 32.25 & 18.27 & 29.70 & 0.6709  & 34.25 & 18.85 & 28.45 & 0.6183  & 85.80 & 21.01 \\
SwinIR+HIRD & 29.73 & 0.8410  & 37.83 & 11.66 & 29.38 & 0.8335  & 42.06 & 12.44 & 29.11 & 0.8255  & 51.47 & 13.60 \\
SwinIR+PWTV & 25.05 & 0.4912  & 766.41 & 37.41 & 23.80 & 0.4505  & 821.29 & 38.79 & 22.15 & 0.3873  & 1415.73 & 39.44 \\
RPNN+HIRD & 33.87 & 0.8931  & 17.47 & 6.90  & 32.75 & 0.8798  & 21.79 & 8.40  & 31.94 & 0.8656  & \textbf{31.09} & \underline{9.86} \\
RPNN+PWTV & 34.91 & 0.8537  & 19.38 & 9.50  & 32.64 & 0.8045  & 24.91 & 11.98 & 30.71 & 0.7371  & 76.43 & 14.97 \\

\midrule

HIRD+PLRD & 28.27 & 0.6445 & 605.39 & 34.60 & 29.18 & 0.6677 & 543.26 & 28.21 & 24.99 & 0.5323 & 561.31 & 41.36 \\
HIRD+SwinIR & 32.28 & 0.8628 & 20.69 & 8.77  & 31.91 & 0.8582 & 24.04 & 9.62  & 31.61 & 0.8500 & 36.18 & 10.81 \\
HIRD+RPNN & 33.11 & 0.9071 & 19.41 & 8.84  & 32.72 & 0.9037 & 22.84 & 9.68  & 32.44 & 0.8970 & 35.26 & 10.88 \\
PWTV+PLRD & 32.42 & 0.8689 & 24.60 & 11.72 & 31.86 & 0.8595 & 27.06 & 13.10 & 31.37 & 0.8512 & 38.51 & 13.67 \\
PWTV+SwinIR & 36.64 & 0.8855 & 14.52 & 5.25  & 35.77 & 0.8749 & 16.83 & 6.01  & 33.42 & 0.8255 & 335.51 & 11.53 \\
PWTV+RPNN & \underline{37.79} & \underline{0.9372} & \underline{12.11} & \underline{4.51} & \underline{37.20} & \underline{0.9336} & \underline{13.71} & \underline{5.18} & \underline{35.41} & \underline{0.9035} & 194.38 & 10.30 \\
\midrule
ZSHipandas (ours) & \textbf{40.24} & \textbf{0.9541} & \textbf{9.07} & \textbf{3.23} & \textbf{38.66} & \textbf{0.9448} & \textbf{10.79} & \textbf{3.91} & \textbf{36.61} & \textbf{0.9227} & 45.26 & \textbf{8.68} \\
\bottomrule
\end{tabular}%
    }
    % \vspace{-1.5em}
  \label{tab:mix}%
\end{table*}%
\begin{figure*}
    \centering
    \includegraphics[width=.95\linewidth]{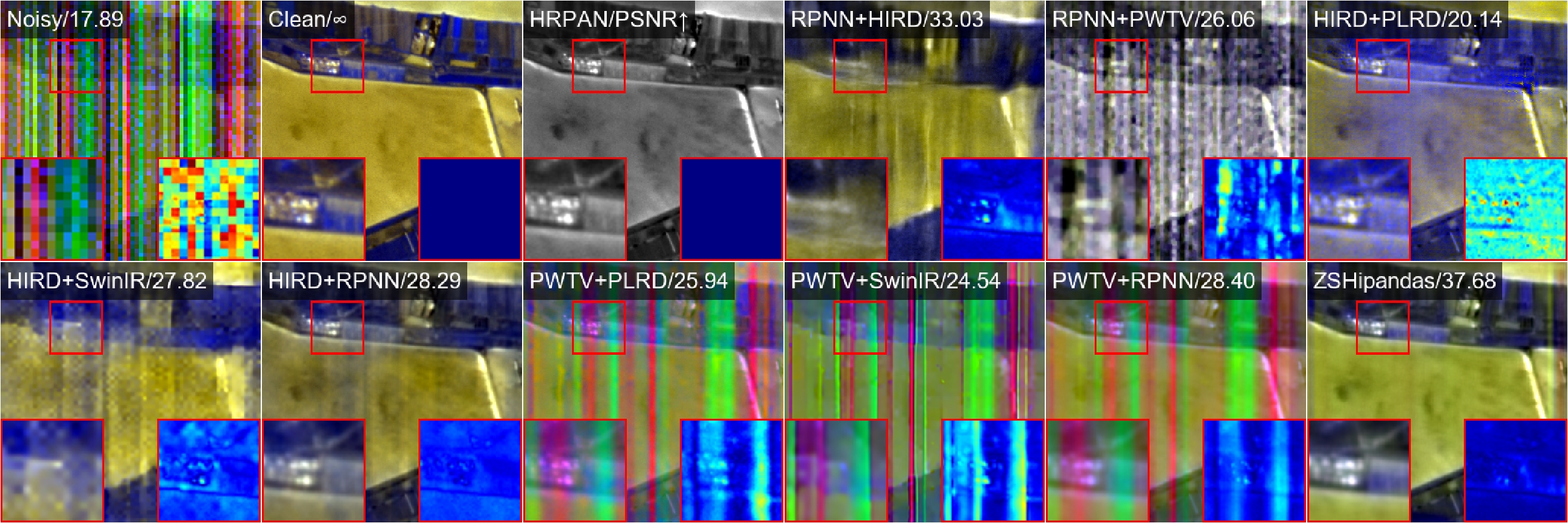}
    \vspace{-.5em}
    \caption{Restoration results for mixture noise with $p=0.55$.}
    \label{fig:mix55}
    \vspace{-.5em}
\end{figure*}
% \subsection{Setup}

\begin{figure*}
    \centering
    \includegraphics[width=.95\linewidth]{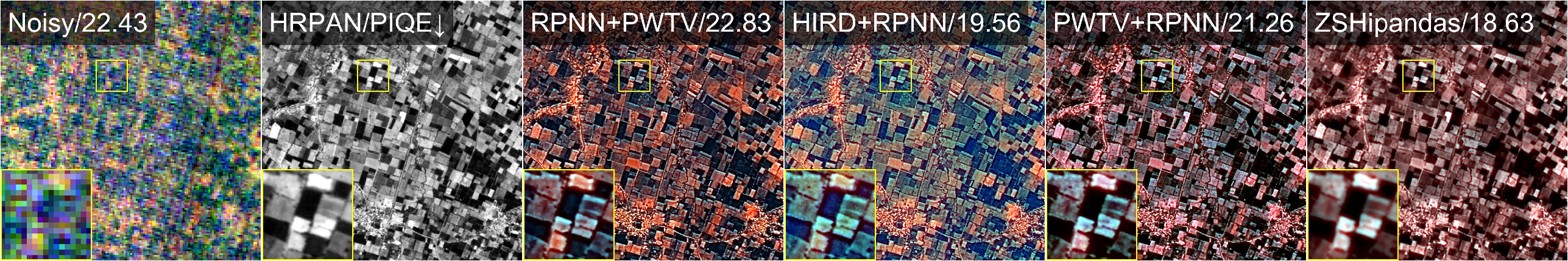}
    \vspace{-.5em}
    \caption{Results on the real-world dataset, Kanpur. PIQE is displayed in the left-upper corner, and lower PIQE indicates better results. }
    \label{fig:realworld}
    \vspace{-.5em}
\end{figure*}

% \vspace{-.5em}
\subsection{Results on simulated datasets}\label{sec:SOTA}
% \vspace{-.5em}
\bfsection{Datasets}
Acquiring paired NLRHS, HRHS and HRPAN images is challenging, so NLRHS and HRPAN images are simulated from HRHS images. A 32-band clean HSI captured over Dongying, Shandong, China, by the Zhuhai-1 satellite, is selected. This image is divided into 81 patches of $256 \times 256$ pixels, which are regarded as HRHS images $H$. The corresponding NLRHS images $N$ are obtained by $N=H\!\downarrow+\varepsilon$, leading to images with a size of $64\times64\times32$. The HRPAN image $P$ is synthesized by applying the spectral response function of the IKONOS satellite to $H$.

\bfsection{Noise configuration} The first group comprises three scenarios: independent and identically distributed (i.i.d.) Gaussian noise with $\sigma=10$ and $30$, and non-i.i.d. Gaussian noise with $\sigma\in[10,50]$. The second group involves mixed noise, where non-i.i.d. Gaussian noise affects 2/3 bands, and impulse, stripe and deadline noise affect 1/3 bands with intensities of $p=0.15$, $0.35$, and $0.55$.

\bfsection{Evaluate metrics} Performance is assessed by peak signal-to-noise ratio (PSNR), structural similarity (SSIM), Erreur Relative Globale Adimensionnelle de Synthese (ERGAS), and spectral angle mapper (SAM). Higher PSNR and SSIM, along with lower ERGAS and SAM, signify better results.

\bfsection{Implement details}
Experiments were conducted on a computer with a RTX 3090 GPU. The training comprised two stages, each consisting of 400 and 600 epochs, with the Adam optimizer of a learning rate $10^{-3}$.
The channel number in networks was fixed at 128. For the GDN and GSRN  components, the respective ranks $r$ were set to 3 and 12.

\bfsection{Compared methods}
ZSHipandas is compared with SOTA methods including
HIRD~\cite{HIRDiff},
PWTV~\cite{PWRCTV},
SwinIR~\cite{SwinIR},
PLRD~\cite{PLRDiff}, and
RPNN~\cite{RPNN}.
The first two are denoising or pandenoising algorithms, and the others are super-resolution or pansharpening algorithms. They are executed in sequence, and two kinds of execution order are considered: super-resolution+denoising, and denoising+super-resolution.

\bfsection{Quantitative Analysis}
\cref{tab:gauss} and \cref{tab:mix} present the evaluation metrics for the Gaussian and mixture noise scenarios, respectively. Compared with PWTV+RPNN (the second best method), ZSHipandas could achieve improvement by 2.3 dB for the case of i.i.d. Gaussian noise with $\sigma=10$, and by 1.3 dB for the case of mixture noise with $p=0.55$. The data illustrate that our proposed ZSHipandas demonstrates a marked enhancement in performance. 

\bfsection{Visual Assessment}
\cref{fig:g30} provides a visual comparison of the restoration results for images corrupted by Gaussian noise at an intensity of $\sigma=30$. It is evident that the images restored by HIRD+SwinIR, PWTV+PLRD, and PWTV+SwinIR still exhibit noticeable noise, and other methods suffer from considerable spectral distortion. Fig. \cref{fig:mix55} depicts the restoration results for images affected by mixture noise at an intensity of $p=0.55$. It reveals that all the compared methods fail to effectively restore the images, with severe noise remaining and significant details being lost. However, the proposed ZSHipandas consistently exhibits good performance across different cases.

\subsection{Results on the real-world dataset}
The methods are applied to a real-world dataset which was captured over Kanpur, India, by the PRISMA satellite. The original dataset is very large, so a small patch is cropped. The NLRHS image is of size $144\times 144\times 49$, and the HRPAN image is of size $864\times 864$, where the ratio of spatial resolution is $s=6$. In this case, RPNN+PWTV, HIRD+RPNN and PWTV+RPNN are carried out,  while many methods are unavailable due to the CUDA out of memory problem. However, these three methods are best performers in simulated experiments, and thus their results are very competitive. 

\cref{fig:realworld} visualizes the results. The local region marked by the box is amplified in the left-bottom corner. It is shown that the image is severely degraded by the noise with a blue color. Restored images by  still preserve the unpleasant color, indicating a spectral distortion. In contrast, ZSHipandas produces a good result with clear textures and satisfactory color. A no-reference metric, Perception based Image Quality Evaluator (PIQE), is reported in \cref{fig:realworld}; and lower PIQE means better quality. The PIQE values also validate our conclusion.

% \vspace{-.5em}
\subsection{Ablation experiments}
% \vspace{-.5em}

Ablation experiments are conducted to validate the rationality of the proposed method. Results are listed in \cref{tab:Ablation} for a noise case, i.i.d. Gaussian noise with $\sigma=10$.

\bfsection{Three components} 
To comprehensively evaluate the effectiveness of each component in ZSHipandas, Exp. I-III respectively remove GDN, GSRN and PRN components by deleting the corresponding loss functions. The results indicate that the removal of any component significantly degrades performance, showing the indispensability.

\bfsection{Two-stage training}
Exp. IV is the scenario without pretraining to validate the rationality of two-stage training. Results show that direct training cannot yield satisfactory performance, suggesting that pretraining is a critical step in preventing the training bias towards noise. 

\bfsection{Low-rank modeling for GDN/GSRN components} 
The architecture of GDN/GSRN components is based on low-rank priors. Exp. V studies the architecture without low-rank prior by revising it as a plain CNN for the fusion of HS and PAN images. The results show that performance is degraded by 3.6 dB without this prior. 

\bfsection{Fusion of PAN images} 
Exp. VI investigates the scenario without PAN images' fusion. It is evident that this case results in substantial degradation, suggesting the proposed model effectively fuses PAN images. 

\begin{table}[t]
    \centering
    \caption{Results of six ablation experiments. \textbf{Bold} indicates the best value.}
    \label{tab:Ablation}
    \vspace{-1em}
    \resizebox{\linewidth}{!}{
\begin{tabular}{cccccc}
\hline
      & Configurations & PSNR  & SSIM  & ERGAS & SAM \bigstrut\\
\hline
I     & w/o $\mathcal{L}_{D}$ & 19.51 & 0.5823 & 57.25 & 19.51 \bigstrut[t]\\
II    & w/o $\mathcal{L}_{S}^{(1)}$ and $\mathcal{L}_{S}^{(2)}$ & 31.49 & 0.9210 & 20.37 & 4.41 \\
III   & w/o $\mathcal{L}_{P}$ and $\mathcal{L}_{Q}$ & 33.32 & 0.7529 & 17.44 & 3.57 \\
IV    & w/o pretraining & 35.94 & 0.9284 & 31.92 & 4.19 \\
V     & w/o low-rank modeling for GDN/GSRN & 36.99 & 0.9325 & 13.35 & 3.83 \\
VI    & w/o fusion of PAN & 33.76 & 0.7706 & 16.91 & 3.54 \bigstrut[b]\\
\hline
      & Ours  & \textbf{40.61} & \textbf{0.9547} & \textbf{8.50} & \textbf{3.00} \bigstrut\\
\hline
\end{tabular}%
        }
    \vspace{-1em}
\end{table}%

\begin{table}[tbp]
  \centering
  \caption{Evaluation metrics between ground truth $L$ and denoised image $\hat{L}$ for the denoising task in the i.i.d. Gaussian noise case with $\sigma=10$.}
  \vspace{-1em}
  \resizebox{0.75\linewidth}{!}{
    \begin{tabular}{ccccc}
    \hline
          & PSNR  & SSIM  & ERGAS & SAM \bigstrut\\
    \hline
    ZSHipandas & \textbf{41.68} & \textbf{0.9610} & \textbf{7.05} & \textbf{2.32} \bigstrut[t]\\
    Only denoising & 40.67 & 0.9522 & 8.60  & 2.93 \bigstrut[b]\\
    \hline
    \end{tabular}%
    }
    \vspace{-1.5em}
  \label{tab:denoising}%
\end{table}%

\subsection{Interaction of pandenoising and pansharpening}
This subsection evaluates the impact of pansharpening on the performance of pandenoising. To address this, we independently train a GDN component with supervision from $\mathcal{L}_{D}$. Subsequently, we compare its denoising capability for LRHS images against ZSHipandas. \cref{tab:denoising} illustrates that employing only the GRN component yields a PSNR value of 40.67. However, when denoising and super-resolution are integrated, as in ZSHipandas, there is a 1 dB enhancement in performance, illustrating the synergy between pandenoising and pansharpening. More analysis can be found in supporting information.

\section{Conclusion}\label{sec:conclusion}
This paper explores the Hipandas problem for HSI joint restoration by image fusion, which is more practical than the sequential implementation of pansharpening and pandenoising. The proposed ZSHipandas is designed by fully exploiting the HSI and detail low-rank priors. With elaborate loss functions and the two-stage training strategy, ZSHipandas outperforms SOTA methods.

{
    \small
    \bibliographystyle{ieeenat_fullname}
    \bibliography{main}
}

\end{document}